\documentclass[11pt]{article}

\usepackage[final]{acl}
\usepackage{multirow}
\usepackage[normalem]{ulem}
\useunder{\uline}{\ul}{}
\usepackage{colortbl} 
\usepackage{times}
\usepackage{amssymb}
\usepackage{latexsym}
\usepackage{CJKutf8}
\usepackage[T1]{fontenc}

\usepackage[utf8]{inputenc}

\usepackage{microtype}

\usepackage{inconsolata}

\usepackage{graphicx}

\title{EVGeoQA: Benchmarking LLMs on Dynamic, Multi-Objective Geo-Spatial Exploration}

\author{
 \textbf{Jianfei Wu\textsuperscript{1}},
 \textbf{Zhichun Wang\textsuperscript{1,2,3}}\thanks{Corresponding author.},
 \textbf{Zhensheng Wang\textsuperscript{1}},
 \textbf{Zhiyu He\textsuperscript{4}},
\\
 \textsuperscript{1}School of Artificial Intelligence, Beijing Normal University, Beijing 100875, China
 \\
 \textsuperscript{2}Beijing Key Laboratory of Artificial Intelligence for Education, Beijing 100875, China
 \\
 \textsuperscript{3}Engineering Research Center of Intelligent Technology and Educational Application,  \\
 Ministry of Education, Beijing 100875, China
 \\
 \textsuperscript{4}College of Computer Science and Technology, National University of Defense Technology,\\
 Changsha 410073, China
\\
 \small{
{\{jianfeiwu, jensenwang\}@mail.bnu.edu.cn, 
{zcwang@bnu.edu.cn},{hezhiyu99@nudt.edu.cn}
 }
}
}

\begin{document}
\maketitle
\begin{abstract}
While Large Language Models (LLMs) demonstrate remarkable reasoning capabilities, their potential for purpose-driven exploration in dynamic geo-spatial environments remains under-investigated. Existing Geo-Spatial Question Answering (GSQA) benchmarks predominantly focus on static retrieval, failing to capture the complexity of real-world planning that involves dynamic user locations and compound constraints. 
To bridge this gap, we introduce EVGeoQA, a novel benchmark built upon Electric Vehicle (EV) charging scenarios that features a distinct location-anchored and dual-objective design. Specifically, each query in EVGeoQA is explicitly bound to a user's real-time coordinate and integrates the dual objectives of a charging necessity and a co-located activity preference.
To systematically assess models in such complex settings, we further propose GeoRover, a general evaluation framework based on a tool-augmented agent architecture to evaluate the LLMs' capacity for dynamic, multi-objective exploration.
Our experiments reveal that while LLMs successfully utilize tools to address sub-tasks, they struggle with long-range spatial exploration. Notably, we observe an emergent capability: LLMs can summarize historical exploration trajectories to enhance exploration efficiency. These findings establish EVGeoQA as a challenging testbed for future geo-spatial intelligence. The dataset and prompts are available at \url{https://github.com/kg-bnu/EVGeoQA}.

\end{abstract}

\section{Introduction}
\begin{figure}[t]
  \includegraphics[width=1\columnwidth]{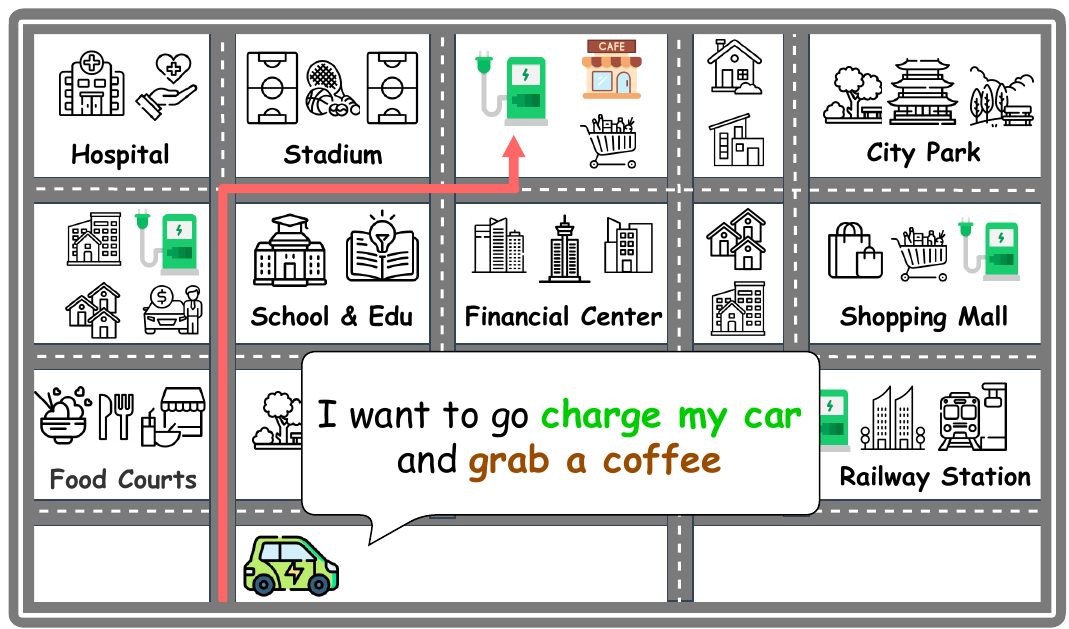}
  \caption{An illustrative EVGeoQA query. Identifying the optimal target requires combining semantic understanding with real-time location and POI information.}
  \label{fig:Task}
\end{figure}
The rapid evolution of LLMs has propelled the development of autonomous agents capable of intricate planning and tool utilization \citep{yao2022react, schick2023toolformer, xi2025rise}. While LLMs excel in processing textual knowledge, grounding them in dynamic geo-spatial environments presents a unique challenge, primarily due to the inherent complexity and immense diversity of realistic spatial scenarios \citep{GeoChallengeMai}. Recent efforts in GSQA have attempted to bridge this gap, yet most existing benchmarks remain limited to static retrieval \citep{feng2023geoqamap, li2025mapqa}. For instance, a typical query might ask, \textit{"What is the distance from the airport to the central railway station?"}. Such tasks rely solely on static spatial topology, neglecting the complexity of real-world mobility where decision-making is constrained by the user's dynamic location and composite demands \citep{Zheng2014UrbanCC}.

The EV charging domain exemplifies this complexity and serves as an ideal yet under-explored testbed. Due to the extended wait times associated with EV charging compared to traditional refueling \citep{Philipsen2018RunningOE}, users are inclined to bundle this service with secondary activities to utilize the duration efficiently. As illustrated in Figure \ref{fig:Task}, a typical inquiry exemplifies this coupled requirement: \textit{"I want to go charge my car and grab a coffee."} Consequently, the optimal solution depends not merely on the charging station itself, but on satisfying a compound objective involving the station's location relative to the user's real-time coordinates and the suitability of the surrounding Point of Interest (POI) context.


To systematically evaluate the geo-spatial reasoning capabilities of LLMs within this rigorous settings, we introduce EVGeoQA, a benchmark built upon the EV charging domain and designed for purpose-driven geo-spatial exploration. Unlike traditional GSQA datasets \citep{li2025mapqa, Geomap1809, Punjani2018TemplateBasedQA}, each query in EVGeoQA is explicitly bound to a user's real-time coordinate and integrates the dual objectives of a charging necessity and a co-located activity preference. Distinguished by this unique design, EVGeoQA shifts the focus from static fact-checking to dynamic planning. 

Our dataset covers three representative Chinese cities—Hangzhou, Qingdao, and Linyi—spanning a hierarchical gradient from major metropolis to developing city. Furthermore, regarding user location generation, Instead of using traditional random coordinate sampling, we propose a synthesis strategy based on K-Means clustering \citep{ahmed2020k} that integrates population density and road network data. By weighting these factors, we simulate user coordinates that statistically align with authentic spatial query distributions, thereby mitigating the spatial bias inherent in random sampling.


With this realistic testbed established, we further propose GeoRover, a general evaluation framework based on a tool-augmented agent architecture to investigate LLMs' geo-spatial exploration capabilities. Specifically, we design four interactive geo-spatial tools to enable the agent to iteratively explore the environment, synthesize historical exploration trajectories, and derive final answers.

Our experiments demonstrate that while accuracy remains relatively high when the answer lies within a short distance, it deteriorates significantly as the exploration distance increases. For instance, the average Hits@2 scores decline from $\sim$50\% to $\sim$38\% as the explore radius expands. This performance degradation underscores the critical limitations of LLMs in long-range spatial exploration. Intriguingly, we observe a spontaneous phenomenon: even in the absence of explicit instructions, LLMs actively summarize historical exploration trajectories to enhance exploration efficiency. These findings establish EVGeoQA as a challenging benchmark for future geo-spatial intelligence.


In summary, our contributions are as follows:
\begin{itemize}
    \item We introduce EVGeoQA, the first GSQA benchmark designed for dynamic multi-objective geo-spatial exploration. It uniquely integrates dynamic user locations with dual-objective constraints to evaluate LLM performance in geo-spatial tasks.
    
    \item We propose GeoRover, a general evaluation framework utilizing a tool-augmented agent equipped with interactive geo-spatial tools to enable active, multi-step exploration, thereby assessing LLM performance in this pervasive yet previously overlooked domain of multi-objective geo-spatial reasoning.
    
    \item Our empirical evaluation reveals that while current LLMs struggle with long-range spatial reasoning, they exhibit a latent ability to summarize historical trajectories, positioning EVGeoQA as a rigorous benchmark for future research.
    
\end{itemize}

\section{Related Work}


\subsection{Benchmarks for GSQA}

The landscape of GSQA has been shaped by foundational benchmarks such as GeoQA201 \citep{Punjani2018TemplateBasedQA}, GeoQA1809 \citep{Geomap1809}, and subsequent works like MapQA \citep{li2025mapqa} and GeoQAMap \citep{feng2023geoqamap}, which predominantly focus on static retrieval over offline databases, neglecting the complexity of real-world environments. Driven by the rapid advancements in Embodied AI, benchmarks such as OpenEQA \citep{majumdar2024openeqa}, SQA3D \citep{masqa3d}, and ScanQA \citep{azuma2022scanqa} have also catalyzed the development of the GSQA domain. However, they are primarily confined to small-scale indoor scenarios with static scene representations.
To effectively address geo-spatial exploration at large scale, LLMs are required to possess synergistic capabilities in planning \citep{xie2024travelplanner, song2023llm}, active exploration \citep{zhou2024navgpt}, and information summarization \citep{chen2023walking, liang2023unleashing}. This multi-faceted requirement significantly escalates the complexity and challenges of the task.

\subsection{Applications of LLMs in the GSQA Domain}
\label{sec:agent}

Driven by exceptional inductive reasoning and information synthesis capabilities, LLMs have been broadly applied to a wide range of real-world tasks, including financial analysis \citep{singh2024finqapt, wang2025financial}, data annotation \citep{wu2025enhancing, wang2024human}, and intelligent education \citep{sun2024llm4edukg, scimkglu2026}. This proficiency in managing complex logical tasks is naturally extending into the geo-spatial domain to address the challenges of interpreting geographical information. 
To handle the unique spatial constraints and heterogeneous data inherent to this field, researchers have integrated LLMs with established autonomous agent frameworks such as ReAct \citep{yao2022react}, Toolformer \citep{schick2023toolformer}, and ToolLLM \citep{qintoolllm}. These integrations empower LLMs to function as autonomous agents capable of independent decision-making and tool-based interaction within geographic environments. Within this context, several specialized works have recently emerged to enhance geo-spatial reasoning capabilities. For instance, Spatial-RAG \citep{yu2025spatial} introduces a spatial retrieval-augmented generation framework that utilizes a dual-retrieval strategy to resolve real-world spatial reasoning questions. CityGPT \citep{feng2025citygpt} focuses on empowering urban-scale spatial cognition by injecting structural knowledge of street networks and urban functional zones into model parameters through specialized instruction tuning. 
Our benchmark is fundamentally built upon these significant advancements and provides a specialized adaptation for GSQA scenarios.


\section{The EVGeoQA Dataset}
\begin{table}[h]
\centering
\begin{tabular}{lccc}
\hline
\textbf{City}      & \textbf{Hangzhou} & \textbf{Qingdao} & \textbf{Linyi} \\ \hline
\textbf{Stations}  & 258               & 165              & 157            \\
\textbf{Locations} & 997               & 995              & 997            \\
\textbf{POIs}      & 25                & 23               & 21             \\
\textbf{QA Pairs}  & 19940             & 14162            & 14416          \\ \hline
\end{tabular}
\caption{Statistics of the EVGeoQA Dataset.}
\label{tab:datasetscale}
\end{table}

\subsection{Problem Formulation}

The core philosophy of this dynamic exploration task is distinct from traditional GSQA benchmarks: it embodies the behavioral pattern of "going to one place to do two things".
Formally, unlike traditional GSQA queries that rely solely on geo-spatial interaction, a query $Q$ in EVGeoQA is explicitly anchored to a user's real-time coordinate $L_u$. The goal is to find a target charging station $S$ that simultaneously satisfies two constraints:
\begin{itemize}
    \item Charging Necessity: The primary task, where the user explicitly requires charging services for their EV.
    \item Co-located Activity: The station must be within a walkable distance to a POI $P$ that fulfills the user's secondary intent.
\end{itemize}

\subsection{Data Acquisition and Pre-processing}
\label{sec:POI}
\begin{figure*}[t] 
  \centering
  \includegraphics[width=\linewidth]{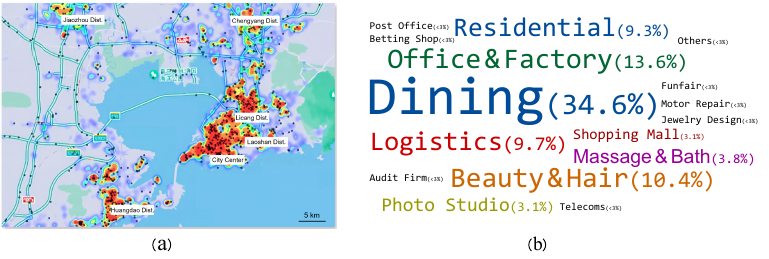}
  \caption{(a) Distribution of query-anchored locations in Qingdao, reflecting the concentration within densely populated regions and along road networks. (b) Distribution of contextual POIs surrounding charging stations in Qingdao, showcasing categorical diversity and non-uniform spatial coverage.}
  \label{fig:DatasetConstruct}
\end{figure*}

To ensure diversity across different urban scales, we selected three representative cities in China: Hangzhou (Provincial Capital), Qingdao (Regional Economic Hub), and Linyi (Prefecture-level City). To construct the geo-spatial foundation for the QA pairs within these regions, we integrate charging station records from the State Grid Corporation of China\footnote{\url{http://www.evs.sgcc.com.cn/}}
 with POI data within a 1km radius of each station retrieved via the Gaode API\footnote{\url{https://lbs.amap.com/}}. Figure \ref{fig:DatasetConstruct}(b) presents the categorical distribution of these contextual POIs.



\subsection{User Location Synthesis via Multi-Source Fusion}

To generate realistic user locations, we synthesize coordinates by leveraging population flow and road network information derived from Baidu heatmap\footnote{\url{https://map.baidu.com/}}, as illustrated in Figure \ref{fig:DatasetConstruct}(a). Formally, we treat the raw heatmap image as a pixel set $\mathcal{X} = \{x_1, x_2, \dots, x_N\}$, where each $x_i \in \mathbb{R}^3$ represents the RGB vector of the $i$-th pixel. We employ the K-Means algorithm \cite{ahmed2020k} to partition these pixels into $K$ semantic clusters $\mathcal{C} = \{C_1, \dots, C_K\}$, representing varying population density tiers and road contours, by minimizing the within-cluster sum of squares:
\begin{equation}
    J = \sum_{k=1}^{K} \sum_{x_i \in C_k} \| x_i - \mu_k \|^2
\end{equation}

where $\mu_k$ denotes the centroid of cluster $C_k$. Subsequently, to simulate realistic human distribution, we assign a density score $w_k$ to each cluster based on its semantic importance. 
The user coordinates are then sampled from these clusters, where the probability $P(k)$ of sampling a location from cluster $C_k$ is determined by a Softmax function.

\begin{equation}
    P(k) = \frac{\exp(w_k )}{\sum_{j=1}^{K} \exp(w_j )}
\end{equation}
The specific weight assignments $w_k$ are detailed in Appendix \ref{app:UserLocationSynthesis}.

\subsection{Dual-Objective Query Generation}
\label{sec:QAGeneration}
Inspired from prior work in template based QA generation \citep{johnson2017clevr, li2025mapqa, wang2025retqa, pampari2018emrqa}, we develop a template-based pipeline to generate natural language queries that reflect the dual-objective nature of the task. 


We first generate structured seed queries by instantiating predefined templates with a refined subset of semantically significant POI categories defined in Section \ref{sec:POI}. We provide more details about this process in Appendix \ref{app:QAGeneration}.

However, raw template-generated seed queries often lack linguistic diversity and purpose-driven context. To address this, we employ a powerful LLM (qwen2.5-72B \citep{qwen2}) equipped with Few-Shot \citep{fewshotlearners} and Chain-of-Thought (CoT) \citep{wei2022chain} prompting techniques to paraphrase these seeds. 
Crucially, this phase involves a functional mapping from static POI categories to realistic intents. For instance, a template slot containing "Stadium" is mapped to activities such as "running" or "exercising". We provide more details about this functional mapping in Appendix \ref{app:paraphrasing}.
\begin{figure*}[h] 
  \centering
  \includegraphics[width=\linewidth]{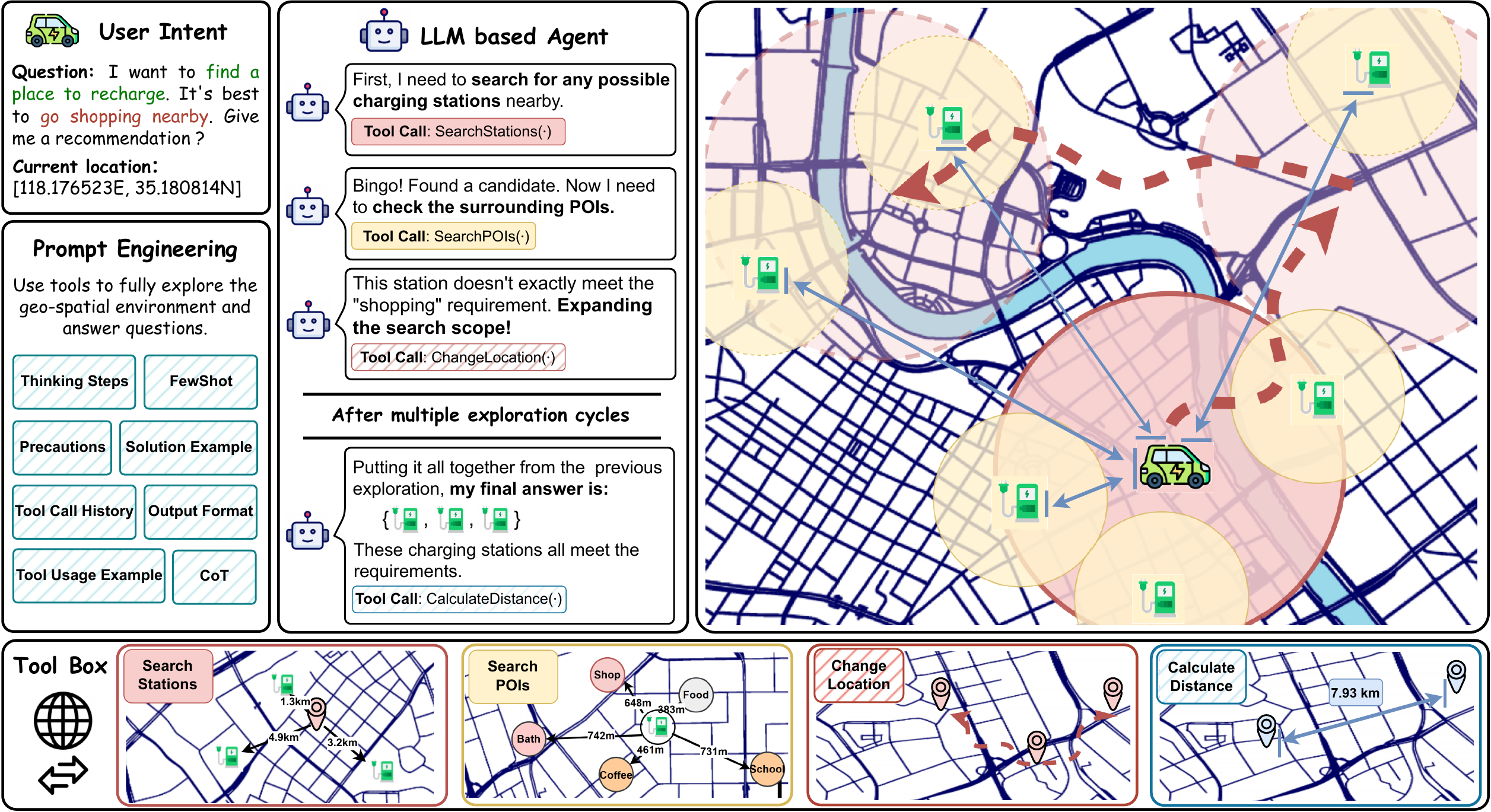}
  \caption{GeoRover Framework Overview. The agent leverages interactive tools to explore the geo-spatial environment, synthesizing the tool invocation trajectory to locate a station that satisfies both charging and activity demands.}
  \label{fig:FrameworkOverview}
\end{figure*}
\subsection{Answer Generation and Quality Control}

To establish ground truth answers, we perform an exhaustive match across all charging station to identify all possible candidates. 
We verify secondary intent alignment by computing cosine similarity between the query's POI slot and station POIs using the CoNAN embedding model\footnote{\url{https://huggingface.co/TencentBAC/Conan-embedding-v2}}, enforcing a stringent threshold of 0.85 to minimize false positives.


Moreover, acknowledging that real-world spatial planning often yields multiple valid optimal solutions,  We rank all semantically valid stations by their vehicle driving distance to the user's query location and retain up to five distinct stations as the ground truth set. 



Finally, we conducted a manual verification of approximately 1000 QA pairs sampled across all POI categories to guarantee the linguistic naturalness and logical correctness of the dataset.

\subsection{Scalability and Representativeness}
Our proposed QA generation pipeline exhibits high extensibility, allowing for seamless adaptation to broader geo-spatial exploration domains beyond the EV charging context. Furthermore, as visualized in Appendix \ref{app:distribution}, the spatial distribution of charging stations spans a wide density spectrum, facilitating the evaluation of both fine-grained discrimination and long-range exploration, thereby ensuring strong universality.

\section{GeoRover Evaluation Framework}
\label{sec:framework}

While EVGeoQA benchmark lays the foundation for multi-objective geo-spatial exploration, effectively evaluating LLMs within this domain necessitates a specialized framework. 
In this section, we introduce GeoRover,  a general evaluation framework based on a tool-augmented agent architecture to systematically investigate the geo-spatial exploration capabilities of LLMs.

\subsection{Geo-Spatial Toolset Definitions}
We prioritize the evaluation of the LLM's geo-spatial exploration capabilities rather than simple information extraction \citep{lewis2020retrieval, ragsurvey}. Consequently, to enforce a realistic setting of partial observability, we design four atomic tools that restrict the agent to obtaining only local information per interaction. This configuration compels the agent to perform iterative reasoning to resolve the query. The specific definitions for these tools are as follows:

\paragraph{SearchStations Tool}
Addressing the user's EV charging demand serves as the foundational step of the resolution process. 
Consequently, the agent is required to explore the environment to identify charging stations near the user's coordinate. 
To enable this retrieval capability, we design the \texttt{SearchStations} tool which allows the agent to perceive the distribution of charging stations within a localized 5km radius centered on a specific coordinate. 

\paragraph{SearchPOIs Tool}
To verify secondary activity constraints, the agent is required to examine the local context of candidate stations to verify their alignment with  specific user demands. The \texttt{SearchPOIs} Tool empowers the agent to inspect the POI context surrounding a specific coordinate, specifically retrieving POIs within a walkable 1km distance.

\paragraph{ChangeLocation Tool}
The \texttt{ChangeLocation} tool functions as the core mechanism of our framework, enabling the agent to actively explore the environment to acquire geographical information from a wider range. Specifically, this tool allows the agent to shift its position from the current coordinate in one of the four cardinal directions by an arbitrary distance. Upon execution, it returns the updated coordinate, which serves as the basis for the agent's subsequent decision-making. By utilizing this new position to re-invoke the \texttt{SearchStations} and \texttt{SearchPOIs} tools, the agent can effectively expand its perception scope, achieving the goal of autonomous spatial exploration.



\paragraph{CalculateDistance Tool}
To facilitate quantitative spatial reasoning, we equip the agent with the \texttt{CalculateDistance} tool. This tool computes the precise vehicle driving distance between coordinates and helps the agent better evaluate the cost-efficiency of candidate targets.

\begin{table*}[!t]
\centering

\fontsize{9}{11}\selectfont 
\resizebox{1\textwidth}{!}{
\begin{tabular}{llccccccccc}
\hline
\textbf{City}                        & \textbf{Distance}                        & \multicolumn{3}{c}{\textbf{\textless 10km}}                                                                     & \multicolumn{3}{c}{\textbf{\textless 20km}}                                                                     & \multicolumn{3}{c}{\textbf{No Limit}}                                                                           \\ \cline{2-11} 
\textbf{}                            & \textbf{Model}                           & \multicolumn{1}{l}{\textbf{Hits@1}} & \multicolumn{1}{l}{\textbf{Hits@2}} & \multicolumn{1}{l}{\textbf{Hits@3}} & \multicolumn{1}{l}{\textbf{Hits@1}} & \multicolumn{1}{l}{\textbf{Hits@2}} & \multicolumn{1}{l}{\textbf{Hits@3}} & \multicolumn{1}{l}{\textbf{Hits@1}} & \multicolumn{1}{l}{\textbf{Hits@2}} & \multicolumn{1}{l}{\textbf{Hits@3}} \\ \hline
                                     & Qwen3-8B                                 & 0.3663                              & 0.4916                              & 0.5381                              & 0.2469                              & 0.3421                              & 0.3867                              & 0.2061                              & 0.2889                              & 0.3295                              \\
                                     & Qwen3-8B*                                & 0.3965                              & 0.4909                              & 0.5011                              & 0.3076                              & 0.3972                              & 0.4513                              & 0.2637                              & 0.3452                              & 0.4375                              \\
                                     & Qwen3-30B-a3b                            & 0.3445                              & 0.5251                              & 0.5810                              & 0.2718                              & 0.3717                              & 0.4012                              & 0.2339                              & 0.2942                              & 0.3133                              \\
                                     & Qwen3-30B-a3b*                           & 0.3906                              & 0.5396                              & 0.5667                              & 0.2804                              & 0.3979                              & 0.4285                              & 0.2283                              & 0.3215                              & 0.3418                              \\
                                     & Qwen25-72b                               & 0.4833                              & \textbf{0.6379}                     & 0.7073                              & 0.3827                              & \textbf{0.5479}                     & 0.6037                              & 0.3333                              & \textbf{0.4878}                     & 0.5614                              \\
                                     & GPT-OSS-20B*                             & 0.3906                              & 0.5322                              & 0.6379                              & 0.3085                              & 0.4302                              & 0.5121                              & 0.2612                              & 0.3762                              & 0.4479                              \\
                                     & GPT-OSS-120B*                            & 0.3417                              & 0.4484                              & 0.4828                              & 0.3075                              & 0.3432                              & 0.3973                              & 0.1842                              & 0.2187                              & 0.2415                              \\
                                     & Gemini-2.5-Flash                         & 0.4648                              & 0.5214                              & 0.6890                              & \textbf{0.4121}                     & 0.4713                              & 0.6147                              & 0.3513                              & 0.4114                              & 0.5110                              \\
                                     & Gemini-2.5-Pro*                          & \textbf{0.4871}                     & 0.5398                              & \textbf{0.7173}                     & 0.4034                              & 0.4812                              & \textbf{0.6373}                     & \textbf{0.3616}                     & 0.4304                              & \textbf{0.5921}                     \\ \cline{2-11} 
\multirow{-10}{*}{\textbf{Hangzhou}} & \cellcolor[HTML]{EFEFEF}\textbf{Average} & \cellcolor[HTML]{EFEFEF}0.4073      & \cellcolor[HTML]{EFEFEF}0.5252      & \cellcolor[HTML]{EFEFEF}0.6024      & \cellcolor[HTML]{EFEFEF}0.3245      & \cellcolor[HTML]{EFEFEF}0.4203      & \cellcolor[HTML]{EFEFEF}0.4925      & \cellcolor[HTML]{EFEFEF}0.2693      & \cellcolor[HTML]{EFEFEF}0.3527      & \cellcolor[HTML]{EFEFEF}0.4196      \\ \hline
                                     & Qwen3-8B                                 & 0.2687                              & 0.4250                              & 0.4875                              & 0.2461                              & 0.3423                              & 0.3807                              & 0.2132                              & 0.2773                              & 0.3161                              \\
                                     & Qwen3-8B*                                & 0.2361                              & 0.3770                              & 0.5245                              & 0.2056                              & 0.3295                              & 0.4318                              & 0.1401                              & 0.2424                              & 0.3129                              \\
                                     & Qwen3-30B-a3b                            & 0.2705                              & 0.4176                              & 0.4664                              & 0.1653                              & 0.2615                              & 0.3012                              & 0.1038                              & 0.1872                              & 0.2755                              \\
                                     & Qwen3-30B-a3b*                           & 0.2236                              & 0.3802                              & 0.4705                              & 0.1891                              & 0.3076                              & 0.3663                              & 0.1235                              & 0.2125                              & 0.2565                              \\
                                     & Qwen25-72b                               & 0.2638                              & \textbf{0.5625}                     & \textbf{0.6206}                     & 0.2286                              & \textbf{0.4555}                     & \textbf{0.4875}                     & 0.1678                              & \textbf{0.3385}                     & \textbf{0.3805}                     \\
                                     & GPT-OSS-20B*                             & 0.2712                              & 0.4133                              & 0.5894                              & 0.2124                              & 0.2996                              & 0.3723                              & 0.1513                              & 0.2012                              & 0.2819                              \\
                                     & GPT-OSS-120B*                            & 0.2254                              & 0.3521                              & 0.4703                              & 0.2014                              & 0.2477                              & 0.3934                              & 0.1433                              & 0.2310                              & 0.3173                              \\
                                     & Gemini-2.5-Flash                         & 0.3073                              & 0.3974                              & 0.4673                              & 0.2857                              & 0.3911                              & 0.4198                              & \textbf{0.2673}                     & 0.3350                              & 0.3789                              \\
                                     & Gemini-2.5-Pro*                          & \textbf{0.3590}                     & 0.4218                              & 0.5383                              & \textbf{0.3106}                     & 0.3583                              & 0.4467                              & 0.2588                              & 0.2738                              & 0.3191                              \\ \cline{2-11} 
\multirow{-10}{*}{\textbf{Qingdao}}  & \cellcolor[HTML]{EFEFEF}\textbf{Average} & \cellcolor[HTML]{EFEFEF}0.2695      & \cellcolor[HTML]{EFEFEF}0.4163      & \cellcolor[HTML]{EFEFEF}0.5150      & \cellcolor[HTML]{EFEFEF}0.2272      & \cellcolor[HTML]{EFEFEF}0.3326      & \cellcolor[HTML]{EFEFEF}0.4000      & \cellcolor[HTML]{EFEFEF}0.1743      & \cellcolor[HTML]{EFEFEF}0.2554      & \cellcolor[HTML]{EFEFEF}0.3154      \\ \hline
                                     & Qwen3-8B                                 & 0.3043                              & 0.4637                              & 0.5362                              & 0.2376                              & 0.3564                              & 0.4257                              & 0.1846                              & 0.2923                              & 0.3461                              \\
                                     & Qwen3-8B*                                & 0.4216                              & \textbf{0.5929}                     & 0.6866                              & 0.3325                              & 0.4828                              & 0.5674                              & 0.3157                              & \textbf{0.4416}                     & \textbf{0.5434}                     \\
                                     & Qwen3-30B-a3b                            & 0.2899                              & 0.4927                              & 0.5507                              & 0.2277                              & 0.3861                              & 0.4455                              & 0.1769                              & 0.3153                              & 0.3615                              \\
                                     & Qwen3-30B-a3b*                           & 0.4512                              & 0.5487                              & 0.6671                              & 0.3703                              & 0.4537                              & 0.5660                              & 0.3253                              & 0.4047                              & 0.5161                              \\
                                     & Qwen25-72b                               & 0.4578                              & 0.5122                              & 0.6582                              & 0.3669                              & 0.4205                              & 0.5773                              & 0.3253                              & 0.3821                              & 0.5357                              \\
                                     & GPT-OSS-20B*                             & 0.3557                              & 0.4496                              & 0.4931                              & 0.2810                              & 0.3987                              & 0.4407                              & 0.2365                              & 0.3494                              & 0.3837                              \\
                                     & GPT-OSS-120B*                            & 0.3653                              & 0.4722                              & 0.4763                              & 0.3121                              & 0.4114                              & 0.4218                              & 0.2411                              & 0.3414                              & 0.3619                              \\
                                     & Gemini-2.5-Flash                         & 0.3993                              & 0.5164                              & 0.5778                              & 0.3333                              & 0.4173                              & 0.4667                              & 0.2962                              & 0.3713                              & 0.4225                              \\
                                     & Gemini-2.5-Pro*                          & \textbf{0.4887}                     & 0.5867                              & \textbf{0.6989}                     & \textbf{0.3975}                     & \textbf{0.4898}                     & \textbf{0.6113}                     & \textbf{0.3827}                     & 0.4404                              & 0.4985                              \\ \cline{2-11} 
\multirow{-10}{*}{\textbf{Linyi}}    & \cellcolor[HTML]{EFEFEF}\textbf{Average} & \cellcolor[HTML]{EFEFEF}0.3926      & \cellcolor[HTML]{EFEFEF}0.5150      & \cellcolor[HTML]{EFEFEF}0.5939      & \cellcolor[HTML]{EFEFEF}0.3177      & \cellcolor[HTML]{EFEFEF}0.4241      & \cellcolor[HTML]{EFEFEF}0.5025      & \cellcolor[HTML]{EFEFEF}0.2760      & \cellcolor[HTML]{EFEFEF}0.3709      & \cellcolor[HTML]{EFEFEF}0.4410      \\ \hline
\end{tabular}
}
\caption{Experimental results on the EVGeoQA dataset, with the best results in \textbf{bold}. LLMs employing the \textit{Thinking} mechanism are marked with *.}
\label{Tab:Mainresult}
\end{table*}

As illustrated in Figure \ref{fig:FrameworkOverview}, it is crucial to emphasize that the invocation sequence and frequency of these four tools are not pre-defined but are dynamically determined by the agent itself. The agent autonomously directs the exploration, synthesizing historical observations to assess information sufficiency and determine when to terminate.

To further enhance the agent's task comprehension and reasoning stability, we incorporate the Few-Shot \citep{fewshotlearners} and Chain-of-Thought (CoT) \citep{wei2022chain} prompting techniques. From an implementation perspective, all APIs are developed upon the Gaode platform \footnote{\url{https://lbs.amap.com/}}. To ensure experimental rigor, we apply strict filtering mechanisms to the raw API returns. This process eliminates irrelevant noise and aligns the retrieved data with the ground truth distributions of the EVGeoQA, thereby maximizing the accuracy of the evaluation results.

\section{Experiment}

\subsection{LLMs Selection}
To comprehensively evaluate the geo-spatial exploration capabilities of LLMs utilizing the EVGeoQA benchmark, we select a diverse set of LLMs representing different parameter scales and reasoning paradigms. Specifically, we include the Qwen series (Qwen3-8B, Qwen3-30B-a3b, and Qwen2.5-72B) \citep{qwen2, qwen3}, the GPT-OSS series (20B and 120B) \citep{openai2025gptoss}, and the Gemini-2.5 family (Flash and Pro) \citep{gemini}. Furthermore, to investigate the specific impact of explicit reasoning on this task, we evaluate the "Thinking" variants for a subset of these LLMs, including Qwen3-8B, Qwen3-30B-a3b, GPT-OSS-20B, GPT-OSS-120B, and Gemini-2.5-Pro. 

\subsection{Evaluation Metrics}
Given the multi-solution nature of real-world spatial planning, we employ Hits@$K$ ($K=1, 2, 3$) as the primary metric to assess the accuracy of the recommended charging stations. Specifically, a prediction is considered a valid 'hit' if it matches any station within the ground truth set. To systematically analyze performance across different spatial scales, we split the dataset by geodesic distance between the user's location and the optimal target station. The samples are divided into three difficulty tiers:

\begin{itemize}
    \item $<10$km (twice the search radius of the \textit{SearchStations} Tool): Scenarios where the target is within a short driving radius.
    \item $<20$km (four times the search radius of the \textit{SearchStations} Tool): Scenarios requiring medium-range planning.
    \item No Limit: The most challenging setting with no distance constraints.
\end{itemize}

\subsection{Main Results and Analysis}
\label{subsec:main_results}

As illustrated in Table \ref{Tab:Mainresult}, while large-scale models perform reasonably well in short-range scenarios, their performance drops significantly as the exploration distance increases. Evidently, there is a clear gap between current model performance and the requirements for reliable geo-spatial agents. We summarize three major findings below. 

\noindent\textbf{LLM "laziness" induces insufficient exploration and performance degradation.} 
There is a consistent and pronounced performance degradation across all LLMs as the exploration radius expands. For instance, in Hangzhou, the average Hits@2 score drops from 0.5252 to 0.3527 as the evaluation shifts from the local scope to the No Limit setting. We attribute this decline primarily to a pervasive 'laziness' phenomenon: when faced with long-range exploration demands, LLMs often prematurely terminate the exploration process. Instead of conducting a sufficient exploration, they tend to fabricate seemingly plausible answers based on limited information retrieved from the previous steps.   

\noindent\textbf{"Thinking" mechanisms enhance exploration depth via retrospective reflection.}
We find that LLMs equipped with explicit "Thinking" modes consistently outperform their standard counterparts. For instance, in the Hangzhou No Limit setting, Qwen3-8B-thinking* achieves a Hits@2 score of 0.3452, showing a distinct advantage over the standard Qwen3-8B (0.2889). We attribute this efficacy to the model's capacity to reflect on historical search trajectories. Unlike standard models that are prone to premature termination, "Thinking" models actively evaluate the sufficiency of retrieved information against the dual-objective constraints and conduct additional exploration steps when information is deemed insufficient. 
We quantitatively analyze this phenomenon in Section \ref{sec:toolusage}.

\noindent\textbf{The Scaling Law persists in Geo-Spatial Exploration Task.} 
Large-scale foundation models, such as Qwen2.5-72B and Gemini-2.5-Pro, consistently dominate across all metrics. In contrast, smaller models (e.g., Qwen3-8B) exhibit a steeper performance decline as task difficulty increases, suggesting that limited parameter counts restrict the capacity to process the high-load spatial contexts inherent to dense urban environments.

\subsection{Analysis of ChangeLocation Tool Usage}
\label{sec:toolusage}

\begin{table}[h]

\fontsize{8}{10}\selectfont
\begin{tabular}{llccc}
\hline
\textbf{City}                    & \textbf{Distance}               & \textbf{\textless 10km}      & \textbf{\textless 20km}      & \textbf{No Limit}            \\ \hline
                                 & Qwen3-8b                        & 0.35                         & 1.36                         & 2.12                         \\
                                 & Qwen3-8b*                       & 0.79                         & 2.11                         & 4.03                         \\
                                 & Qwen3-30b-a3b                   & 0.37                         & 1.13                         & 3.26                         \\
                                 & Qwen3-30b-a3b*                  & 0.42                         & 1.72                         & 3.74                         \\
                                 & Gemini-2.5-Flash                & 0.40                         & 2.77                         & 3.91                         \\
                                 & Gemini-2.5-Pro*                 & 1.12                         & 3.31                         & 5.62                         \\ \cline{2-5} 
\multirow{-7}{*}{\textbf{Linyi}} & \cellcolor[HTML]{EFEFEF}Average & \cellcolor[HTML]{EFEFEF}0.58 & \cellcolor[HTML]{EFEFEF}2.07 & \cellcolor[HTML]{EFEFEF}3.78 \\ \hline
\end{tabular}
\caption{Average tool invocation frequency of the \texttt{ChangeLocation} Tool across different distance tiers in Linyi.}
\label{tab:tool_frequency}
\end{table}


As discussed in Section \ref{sec:framework}, the \texttt{ChangeLocation} tool is the core mechanism for expanding exploration scope. To quantitatively assess how LLMs utilize this capability, we record its average invocation frequency across different difficulty tiers in Linyi. Specifically, this metric is defined as the mean number of times the \texttt{ChangeLocation} tool is invoked by the agent within a single exploration episode.

The results in Table \ref{tab:tool_frequency} reveal that the invocation frequency is much lower than anticipated, especially on long-range tasks. We attribute this to two primary factors. First, consistent with the "laziness" bottleneck, agents often terminate exploration prematurely without exhaustively exploring the environment. Second, we observe an emergent capability in advanced models to synthesize spatial contexts from interaction history and infer new coordinates without explicitly invoking the tool.


Despite these nuances, a distinct positive correlation exists between the frequency of tool usage and the agent's performance in complex scenarios. This increased active exploration directly aligns with the superior accuracy reported in our main results in Table \ref{Tab:Mainresult}, confirming that active exploration is a determinant factor for success in large-scale geo-spatial planning tasks.
\begin{figure}[h] 
  \centering
  \includegraphics[width=\linewidth]{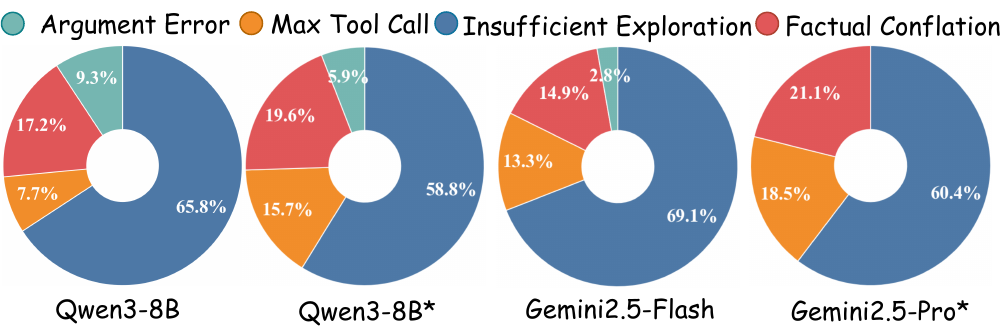}
  \caption{Distribution of Error Causes in Linyi}
  \label{fig:error}
\end{figure}
\begin{figure*}[!t] 
  \centering
  \includegraphics[width=\linewidth]{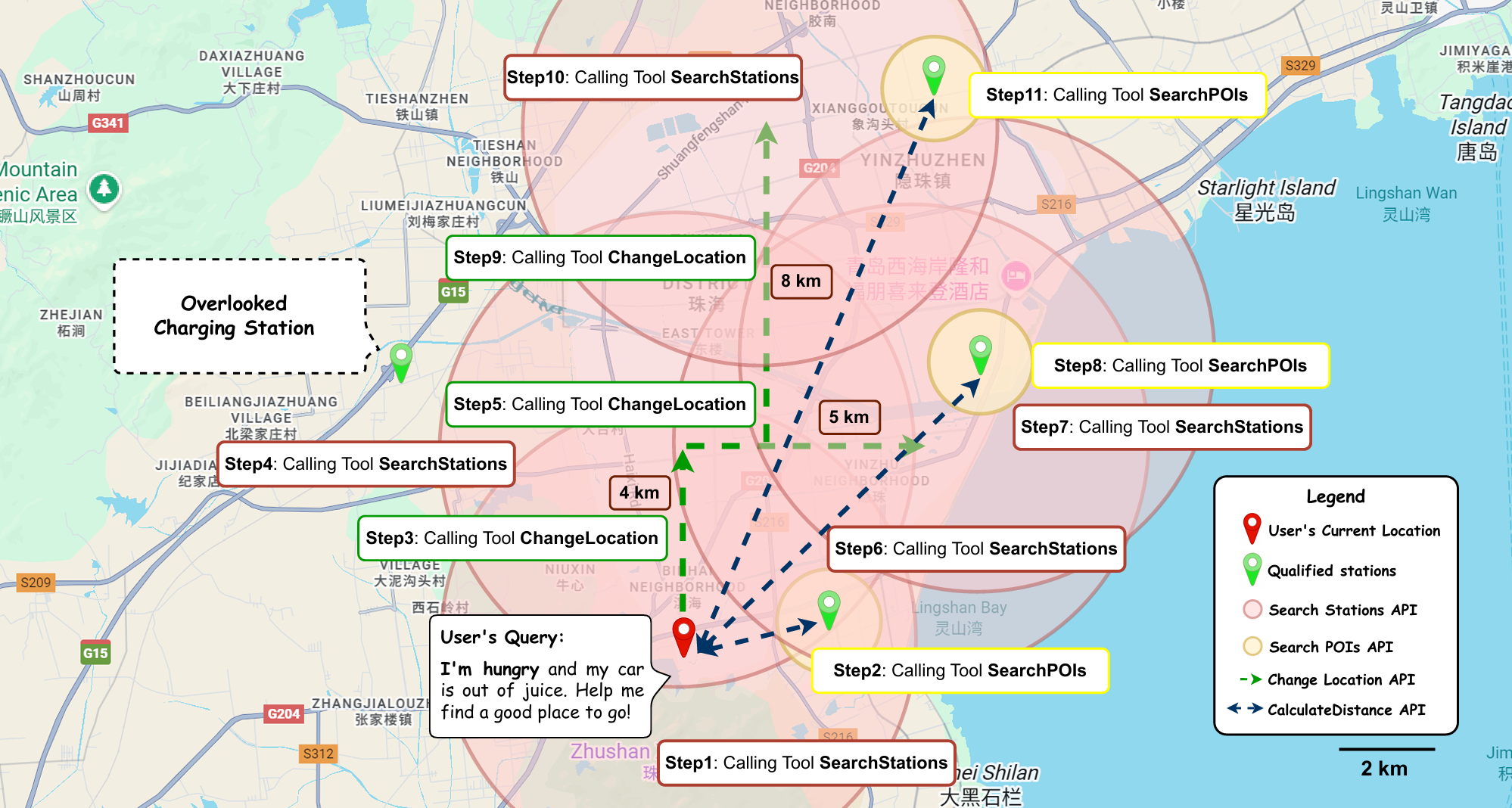}
  \caption{Case Study: Visualization of a multi-step exploration trajectory by Gemini-2.5-Pro* in Qingdao.}
  \label{fig:CaseStudy}
\end{figure*}
\subsection{Error Analysis}

As illustrated in Figure \ref{fig:error}, a significant portion of failures stems from the insufficient exploration across the environment. All LLMs-based agents exhibit a substantial degree of "laziness" when facing complex planning tasks. This suggests that current LLMs lack the capacity to maintain a long-horizon search strategy without explicit guidance or reinforcement.

Furthermore, we observe that the integration of heterogeneous data, such as exploration trajectories, charging stations and POI details, triggers the "Lost in the Middle" phenomenon \citep{liu2024lost, lilong}. Agents frequently conflate attributes in these complex long contexts, producing responses that are linguistically fluent but factually erroneous.

Finally, Except for Gemini2.5-Pro, all evaluated LLMs encounter argument-related errors to varying degrees, indicating that the effective invocation of geo-spatial tools remains a significant challenge. 

The detailed definitions and settings for these error categories are provided in Appendix \ref{sec:errorinfo}.

\subsection{Case Study}
\label{sec:case_study}

We also qualitatively examine how agents navigate and reason within these complex geo-spatial environments. As shown in Figure \ref{fig:CaseStudy}, high-performing LLMs can actively summarize historical exploration trajectories to optimize future search steps.


Specifically, the agent initially executes a localized exploration surrounding the user's query coordinate (Steps 1--4). Upon determining that the nearby area lacks charging stations that satisfy the dual constraints, the agent autonomously performs long-range spatial transfers (Step 5 and 9) to its exploration scope. Although our prompt design provides no predefined exploration rules, the agent appears to synthesize its historical  trajectory, selecting new anchor points that maximize spatial coverage and avoid redundant search efforts. Finally, the agent leverages the \texttt{CalculateDistance} tool to acquire precise distance metrics, synthesizing all accumulated observations to formulate the final recommendation.

This emergent behavior highlights the potential of LLMs to comprehend spatial layouts and conduct purpose-driven exploration. 
However, the agent's exploration process exhibits distinct limitations. As observed in the left quadrant of Figure \ref{fig:CaseStudy}, a qualified charging station located closer to the initial position is overlooked during the exploration process. This omission aligns with the "laziness" phenomenon discussed in Section \ref{subsec:main_results}. This suggests that while LLMs exhibit potential spatial reasoning, their ability to guarantee global optimality in geo-spatial exploration remains a critical bottleneck requiring future optimization.

\section{Conclusion and Discussion}

In this paper, we introduced EVGeoQA, the first benchmark designed to evaluate the purpose-driven exploration capabilities of LLMs within dynamic geo-spatial environments. To facilitate systematic assessment, We further propose GeoRover, a general evaluation framework adopting a tool-augmented agent architecture to investigate LLMs' geo-spatial exploration capabilities.

Our experimental results reveal that while LLMs perform effectively in localized, short-range scenarios, they suffer from pronounced performance degradation in long-range tasks. Although a clear gap exists between current model performance and the requirements for reliable geo-spatial agents , the latent ability of trajectory summarization highlights a significant potential for LLM-based complex geo-spatial reasoning.

By exposing key bottlenecks such as insufficient exploration and attribute conflation, EVGeoQA serves as a rigorous testbed to guide the development of more robust and strategically-aware geo-spatial agents for open-world applications.

\section*{Limitations}

The EVGeoQA benchmark is constructed based on urban data from three distinct Chinese cities of varying scales, with QA pairs predominantly in Chinese. This linguistic specificity may introduce inherent biases. Our future work aims to mitigate this by incorporating multi-lingual data and a broader range of global cities. 

Furthermore, our current evaluation primarily assesses the inherent reasoning capabilities of LLMs. We acknowledge that advanced techniques, such as domain-specific fine-tuning (SFT) \citep{gururangan2020don, zheng2024fine,hu2022lora}, hold significant potential for enhancing model performance in specialized spatial tasks, and investigating the feasibility and efficacy of such optimization strategies constitutes a key direction for our subsequent research. Moving forward, we remain dedicated to the field of geo-spatial exploration, striving to refine our benchmarks and frameworks to further advance embodied spatial intelligence.

\section*{Acknowledgments}
This work was supported by the National Natural Science Foundation of China (No. 62276026) and the Fundamental Research Funds for the Central Universities (No. 2253500001)


\bibliography{main}

\appendix

\section{Appendix}
\label{sec:appendix}
\subsection{User Location Synthesis via Multi-Source Fusion}
\label{app:UserLocationSynthesis}
\begin{figure}[h] 
  \centering
  \includegraphics[width=\linewidth]{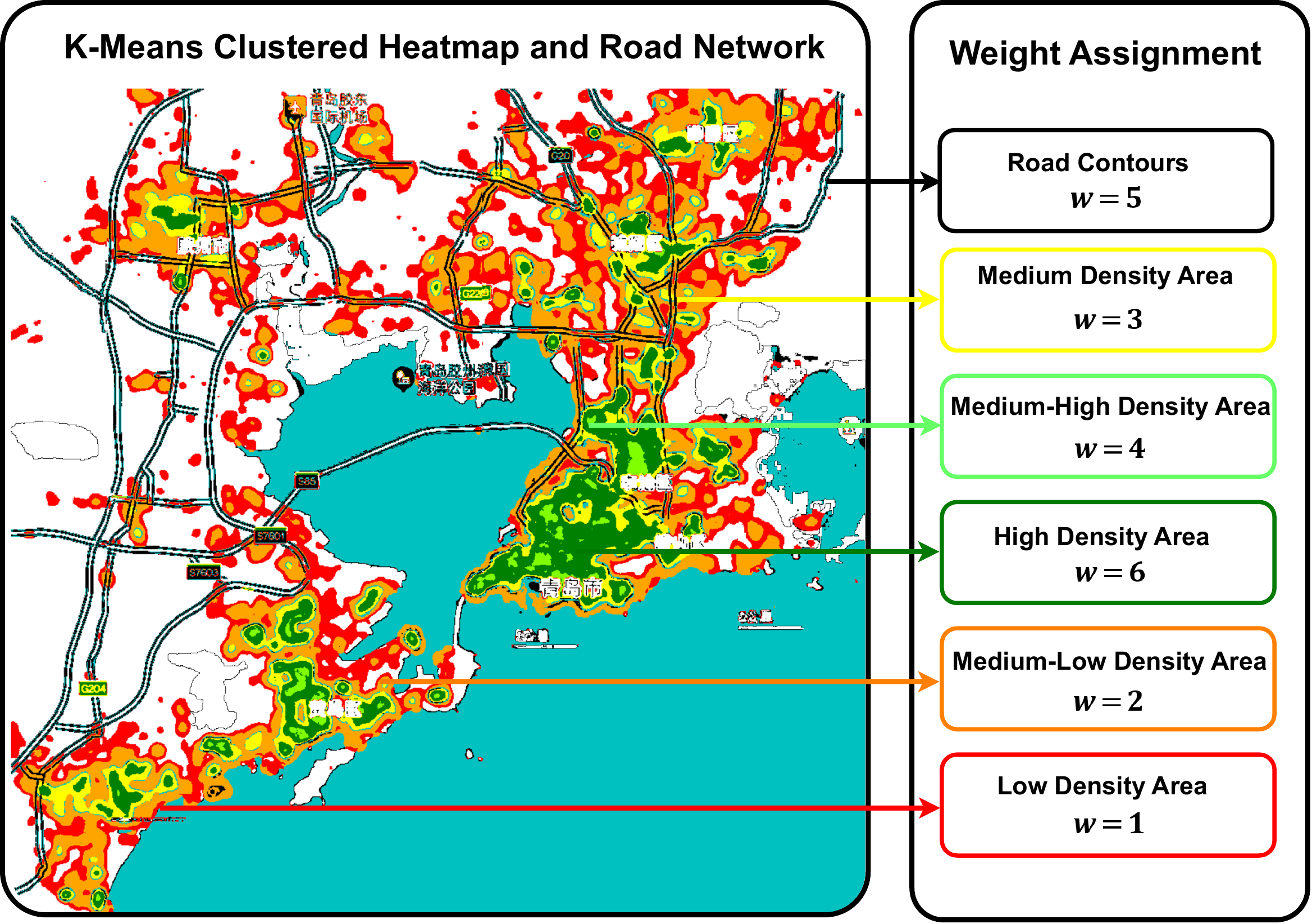}
  \caption{Illustration of the Multi-Source Fusion strategy for user location synthesis.}
\end{figure}

 The left panel displays the spatial segmentation of density tiers and road contours derived via K-Means clustering (exemplified in Qingdao). The right panel details the semantic weight assignment ($w$), where higher weights are explicitly allocated to road networks and high-density regions to prioritize their selection. These weights drive a non-uniform sampling mechanism to generate realistic user coordinates, which subsequently serve as spatial anchors for question generation.

\subsection{Template Based Dual-Objective Query Generation}
\label{Template Based Dual-Objective Query Generation}
\label{app:QAGeneration}
In this section, we present the templates used for generating QA pairs for the EV charging scenario. The placeholder \texttt{\{position\}} denotes the Point of Interest (POI) category in the vicinity of the charging station (e.g., "shopping mall", "park". The templates are presented in their \textbf{original Chinese format} as follows:

\begin{CJK*}{UTF8}{gbsn}
\begin{itemize}

    \item 帮我找一个在 \{position\} 附近的充电桩.
    
    \item 我想去 \{position\} ,并顺便给我的电动汽车充电. 你能不能给我推荐一个充电桩?
    
    \item 我要去 \{position\} 办点事，顺便想给我的电动车充电。哪里比较合适？
    
    \item 有在 \{position\} 附近的充电桩吗？
    
    \item 你能把 \{position\} 区域的充电桩位置给我展示一下吗？
    
    \item 在 \{position\} 附近，有推荐的充电桩吗？
    
    \item 帮我推荐一个距离 \{position\} 最近的充电桩？

\end{itemize}
\end{CJK*}

We also provide the \textbf{English translations} of these templates for reference:

\begin{itemize}

    \item Help me find a charging station near \{position\}.
    
    \item I plan to go to \{position\} and charge my EV. Can you recommend a charging station?
    
    \item I am going to \{position\} to run some errands and want to charge my electric vehicle. Where would be a suitable place?
    
    \item Are there any charging stations near \{position\}?
    
    \item Can you show me the locations of charging stations in the \{position\} area?
    
    \item Do you have any recommended charging stations near \{position\}?
    
    \item Please recommend the charging station closest to \{position\}.

\end{itemize}

Although these seed templates appear relatively simple and exhibit a degree of linguistic redundancy, their combination with the advanced LLM-based paraphrasing strategy (as detailed in Section \ref{sec:QAGeneration}) and a diverse spectrum of POI categories ensures a comprehensive coverage of user query variations within the EV charging domain.


    
    
\subsection{Spatial Distribution of Charging Stations}
\label{app:distribution}
\begin{figure*}[t] 
  \centering
  \includegraphics[width=\linewidth]{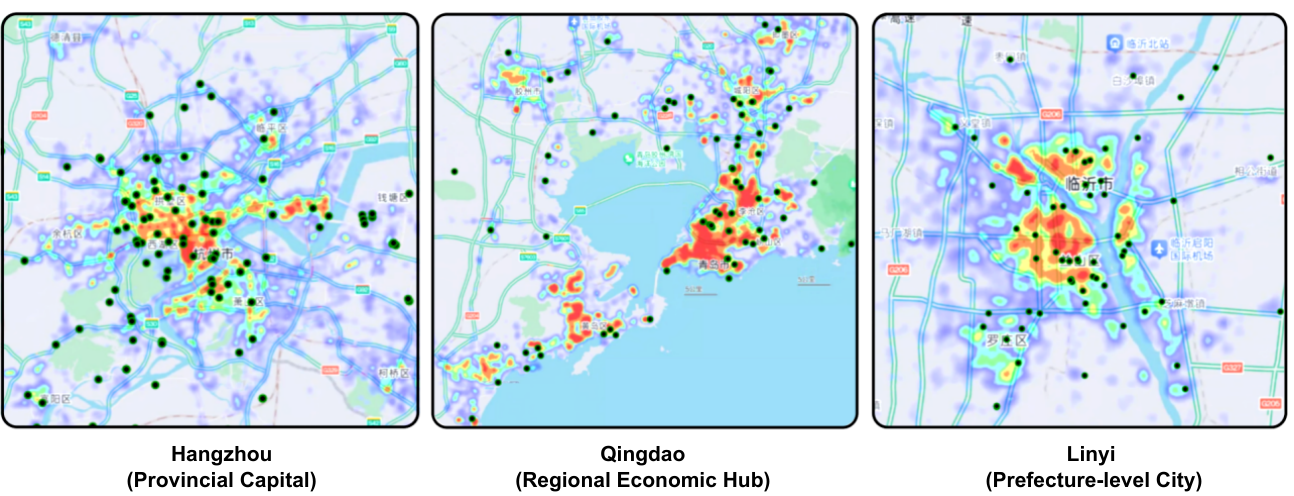}
  \caption{Spatial distribution of charging stations across the three representative cities. Visualizations are zoomed into city centers for clarity. The full dataset is available in our code repository.}
  \label{app:dist}
\end{figure*}
To demonstrate the topological diversity of the EVGeoQA benchmark, we visualize the spatial density of charging stations across the three selected cities: Hangzhou, Qingdao, and Linyi. As illustrated in Figure \ref{app:dist}, the distribution of charging infrastructure exhibits a strong correlation with regional economic development and population density.

Specifically, the distribution patterns reveal two distinct spatial characteristics that correspond to different challenges in geo-spatial reasoning:
\begin{itemize}
    \item \textbf{High-Density Urban Cores:} In economically developed city centers and densely populated districts, charging stations are clustered with high density. In these scenarios,  the agent must filter through numerous candidate stations in close proximity to identify the specific one that optimally satisfies the user's secondary activity constraint (e.g., determining which of the five nearby stations is closest to a specific type of restaurant).
    
    \item \textbf{Sparse Peripheral Regions:} Conversely, in suburban areas and developing districts, the distribution of stations becomes significantly sparser. These scenarios simulate long-range exploration tasks, where the agent cannot rely on immediate availability. Instead, it must engage in multi-step planning and active map traversal to locate feasible resources.
\end{itemize}

\subsection{Definitions and Settings for
Error Analysis}
\label{sec:errorinfo}
We categorize the observed errors into four distinct types. Detailed definitions and their causes are provided below:
\begin{itemize} 

\item \textbf{Argument Error}: This error occurs when an agent generates a response in an incorrect format that violates the tool's API format. 
\item \textbf{Max Tool Call}: This category refers to exploration process that terminate because the agent reaches the maximum limit of 20 tool invocations. This phenomenon typically arises from two scenarios: (1) the model falls into a logical loop by repeatedly invoking the same tool with identical parameters, or (2) the exploration strategy is inefficient, causing the agent to exhaust its call budget before identifying a valid target.

\item \textbf{Insufficient Exploration}: This failure is characterized by premature termination of the search process. Despite failing to locate a charging station that satisfies all constraints during the exploration, the agent ceases its exploration and provides a final incorrect response. This aligns with the "laziness" bottleneck discussed in our primary analysis.

\item \textbf{Factual Conflation}: This error is primarily driven by hallucinations or the "Lost in the Middle" effect. In these cases, although all necessary information to resolve the query has been retrieved during the exploration steps, the agent fails to correctly synthesize the context, leading to a factually erroneous output despite the availability of valid evidence. 
\end{itemize}

\subsection{LLM-based Query Paraphrasing and Functional Mapping} \label{app:paraphrasing}

In real-world geo-spatial QA scenarios, user queries typically exhibit high linguistic variability and implicit intent. Unlike structured seed queries, users rarely specify static POI categories (e.g., \textit{"Find a charging station near a restaurant"}); instead, they tend to express functional needs or activities (e.g., \textit{"I need to charge my car while having dinner"}).

To bridge the semantic gap between template-generated seeds queries and authentic user inquiries, we implement a functional mapping strategy. Specifically, we employ Qwen2.5-72B, a large language model with strong instruction-following capabilities, to paraphrase the initial queries. The rewriting process is governed by three key principles: 
\begin{enumerate} 
\item \textbf{Intent Transformation:} Converting explicit location types into functional descriptions (e.g., mapping "Gym" to "working out"). 
\item \textbf{Contextual Logic:} Ensuring the activity is logically compatible with the charging duration and location type. 
\item \textbf{Linguistic Diversity:} Varying sentence structures and tones to mimic casual, spoken language. \end{enumerate}

To vividly illustrate this transformation, we alse provide three representative examples comparing the raw template-generated seeds queries with their LLM-polished counterparts:

\begin{itemize} \item \textbf{Case 1 (Dining):} \\ \textit{Template:} "Help me find a charging station near \{Dinning/Restaurant\}" \\ \textit{Polished:} "I'm starving and my car is running low. Is there a place where I can eat and charge simultaneously?"

\item \textbf{Case 2 (Exercise):} \\
\textit{Template:} "I plan to go to \{Stadium/Sports Venue\} and charge my EV. Can you recommend a charging station? " \\
\textit{Polished:} "I want to get a workout in while waiting for my EV to charge. Can you recommend a spot?"

\item \textbf{Case 3 (Accommodation):} \\
\textit{Template:} "Do you have any recommended charging stations near \{Hotel/Accommodation\}? \\
\textit{Polished:} "I'm looking for a place to rest for the night, and my car needs power too. Can you help me find a suitable location?"
\end{itemize}

\end{document}